\documentclass{article}

\usepackage{arxiv}

\usepackage[utf8]{inputenc} 
\usepackage[T1]{fontenc}    
\usepackage{hyperref}       
\usepackage{url}            
\usepackage{booktabs}       
\usepackage{amsfonts}       
\usepackage{nicefrac}       
\usepackage{microtype}      
\usepackage{lipsum}		
\usepackage{graphicx}
\usepackage[numbers]{natbib}
\usepackage{doi}

\usepackage{blindtext} 
\usepackage{epigraph}
\usepackage{subfig}
\usepackage{amsmath}
\newcommand\numberthis{\addtocounter{equation}{1}\tag{\theequation}}
\usepackage{xcolor}
\usepackage{listings}
\usepackage{xparse}

\newcommand{\chapquote}[3]{\begin{quotation} \textit{#1} \end{quotation} \begin{flushright} - #2\end{flushright} }
\NewDocumentCommand{\codeword}{v}{%
\texttt{\textcolor{blue}{#1}}%
}

\title{The Study of Complex Human Locomotion Behaviors: From Crawling to Walking}

\author{ \href{https://orcid.org/0000-0003-2335-949X}{\includegraphics[scale=0.06]{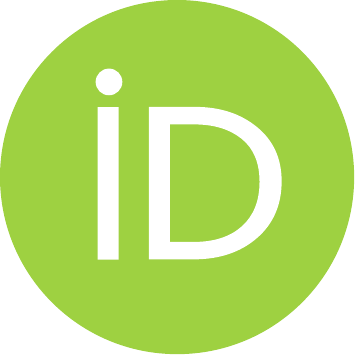}\hspace{1mm}Shengjie Xu} \\
	Department of Electrical and Computer Engineering\\
	University of California, Santa Cruz\\
	Santa Cruz, CA 95064 \\
	\texttt{sxu88@ucsc.edu} \\
	\And
	\href{https://orcid.org/0000-0000-0000-0000}{\includegraphics[scale=0.06]{orcid.pdf}\hspace{1mm}Kevin Mok} \\
	Department of Electrical and Computer Engineering\\
	University of California, Santa Cruz\\
	Santa Cruz, CA 95064 \\
	\texttt{kemok@ucsc.edu} \\
}

\renewcommand{\shorttitle}{\textit{arXiv} Template}

\hypersetup{
pdftitle={The Study of Complex Human Locomotion Behaviors: From Crawling to Walking},
pdfauthor={Shengjie Xu, Kevin Mok},
pdfkeywords={Infant, Locomotion, Mujoco, Simulation, Crawl, Walk, Robotics},
}

\begin{document}
\maketitle

\begin{abstract}
	This paper uses a simple state machine to develop a control algorithm for controlling an infant humanoid in the context of a simple model system. The algorithm is inspired by a baby who starts learning to stand and walk at 7 to 12 months of age: he or she initially learns to crawl and then, once the lower limb muscles are strong enough, can learn to walk by coming to support his or her upper trunk. Ideally, this algorithm-supported locomotion can take the baby to any desired location: a pile of toys, a tasty snack, or the baby's parents or relatives. In this paper we analyze the crawling stage, the simple 2d bipedal model, and the initial walking form from 8 to 18 months of age, and quantitatively evaluate the ideal kinematics model and simulation results for these stages.
\end{abstract}


\section{Introduction}

\chapquote{``Instead of trying to produce a programme to simulate the adult mind, why not rather try to produce one which simulates the child’s? If this were then subjected to an appropriate course of education one would obtain the adult brain."}{Alan Turing \citep{turing1950mind}}


Over the last decades, the growing research in robotics has diverged into three different areas: robot manipulators, mobile robots, and biologically inspired robots \citep{garcia2007evolution}. In the early stages, robots were designed to perform industrial tasks such as welding, painting, and palletizing. With the thrust of military, rescue, and planetary exploration, the research of autonomous mobile-robot endows the navigation ability of robotics. Such navigation algorithm consists of the perception of the environment, localization, motion planning, and motion generation. At this point, the robots have enough intelligence to interact with their surroundings to solve challenging tasks, such as the self-driving challenge or DARPA Robotics Challenge \cite{thrun2006stanley,krotkov2017darpa}. Aside from traditional mobile robotics, there is a lot of interest in using biological inspiration to create new types of robots with adaptive locomotion systems. Some research groups focus on other types of locomotion, such as snake and fish systems. However, we'll focus on humanoid robots, especially baby models, due to their widespread use.

Can a robot learn like a child? How do baby learn to walk? The motivation behind this paper is dominated by the goal of producing robotics that can learn to walk like a baby. Unlike classical robotics, a promising research field in robotics called learning-based robotics is a multidisciplinary field where robotics, computer science, and psychology all meet \citep{adolph2012you}.

In this paper, we modeled two classical walking and running models, based on which we have proposed a quadrupedal model, as our baby crawling model, and simulated the dynamics in the physics engine. We hope that this research will provide a simulation and experimental environment for interdisciplinary research in areas such as computer science, robotics, and psychology.





\section{Formulation of the problem}
\subsection{The Simple Models of Running and Walking}
The intuitive locomotion model we observed from human walking and running gaits can be used to abstract conceptual models of legged locomotion. We can simplify the whole-body locomotion into a simple model that contains the feet, hip, and top of the spine by using tracking marks. We discussed two well-established models, the \textit{template} models of human locomotion, that formalizing the locomotion of running and walking \citep[p. 62]{sharbafi2017bioinspired}.

\label{sec:2D_Bipedal}
\subsubsection{The Spring-Loaded Inverted Pendulum Running Model}
By observing the consecutive running photos of humans, Sharbafi et al. \citep[p. 59]{sharbafi2017bioinspired} have stated that the overall movement of running was constituted by three anatomical parts: trunk, stance leg, and swing leg. The two legs switch functionality by repeating the entire gait cycle: the legs are either posed in stance (with foot colliding with the ground) or in swing (with foot departing the ground), with flight phases occurring when both legs are in the air. This simple ``switching'' philosophy still applies in the state machine we applied in the walking model section.

\begin{figure}[h!]
	\centering
    \includegraphics[width=10cm]{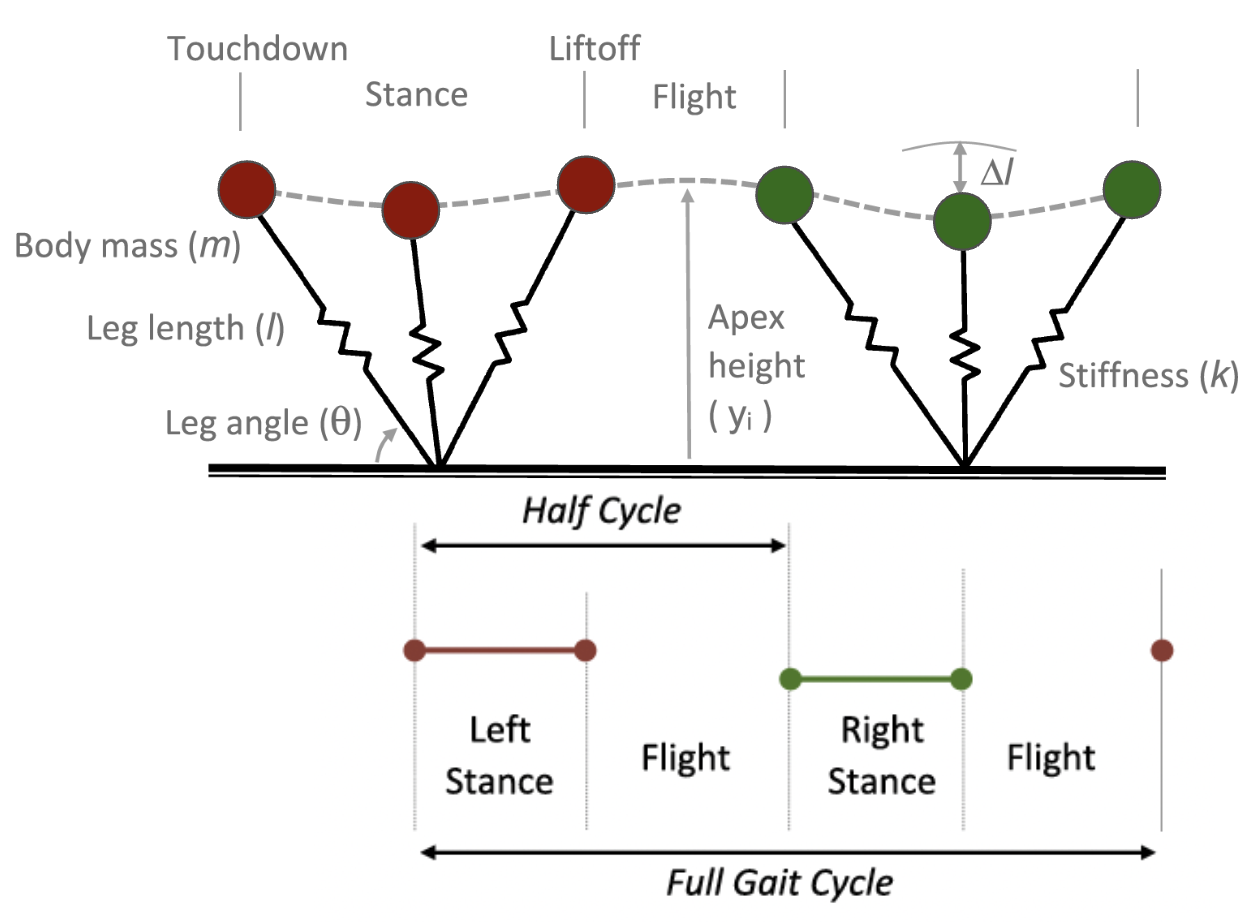}
	\caption{An illustration of the Spring-Loaded Inverted Pendulum (SLIP) model describe a full cycle of running gait. Left leg marked as red, and right leg marked as green \citep[p. 80]{sharbafi2017bioinspired}}.
	\label{fig:Running}
\end{figure}

Blickhan et al.'s \citep{blickhan1993similarity} research on ground reaction force reveals that the leg force is proportional to the leg length in a Hooke's Law way. Alexander \citep{alexander1990three} describes the springs behaviour of legged locomotion as pop stick principle. The recirculating cycle of the swing leg can be observed as a combination of swinging of a passive pendulum and an active, controlling the placement of the future stance leg, \textit{retraction}. The retraction happens at the beginning and the end of the swing phase. Blickhan \citep{blickhan1989spring} describes this overall behavior of running as a spring-loaded inverted pendulum (SLIP) model. As shown in Figure \ref{fig:Running}, SLIP models are low-dimensional models that commonly use a single point-mass as the trunk and a single massless leg to represent the recirculating stance or swing leg.



\subsubsection{Inverted Pendulum Model of Walking}


The overall motions of the trunk, stance leg, and swing leg of walking humans, walking mechanisms and robots, and the inverted pendulum model share many similarities: During the stance phase of human walking, when a single leg is on the ground, the body tends to rise and then fall as it pivots about the foot. This is similar to the way an inverted pendulum moves about its pivot. Walking is also described as a pattern or gait with alternating left and right legs. This characteristic walking pattern can be exhibited by bipedal inverted pendulum model.

\begin{figure}[h!]
	\centering
    \includegraphics[width=8cm]{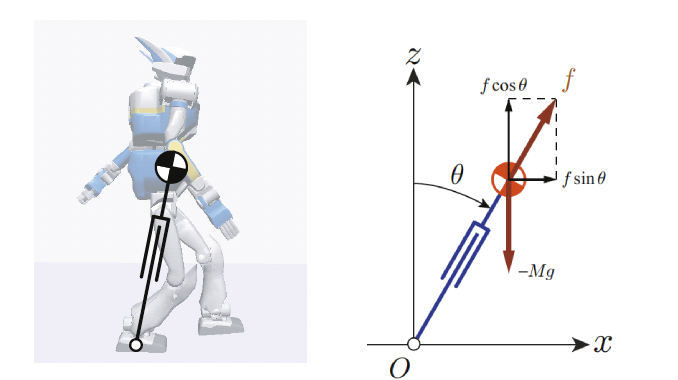}
	\caption{Free-body diagram of a simple Inverted Pendulum model \citep{kajita2014introduction}.}
	\label{fig:2D-Bip}
\end{figure}

The 2D inverted pendulum model is the simplest model for a walking humanoid. As shown in Figure. \ref{fig:2D-Bip}, the IP model consists of the center of mass (CoM) and an ideal swing leg, whose mass is assumed to be negligible. By applying laws of physics, Newton’s Second Law of Mechanics or the Euler–Lagrange Equation, the dynamics of the model is governed by a couple of differential equations as follows: \citep{kajita2014introduction,sharbafi2017bioinspired}

\begin{align*}
    r^2\ddot{\theta}+2r\dot{r}\dot{\theta}-gr\sin{\theta}&=\tau/M\\
    \ddot{r}-r\dot{\theta}^2+g\cos{\theta}&=f/M\\
\end{align*}

, where $M$ is the trunk mass, $r$ is the leg length, $g$ is gravity, $\theta$ is the inclination angle of the pendulum, $\theta$ is the angle of the inclined plane, $\tau$ is the torque at the pivot, and $f$ is the force at the prismatic joint along the leg.

The horizontal dynamic of CoM is noteworthy to consider as initiated by a kick force:


\begin{equation}\label{eqn_2}
    f=\frac{Mg}{\cos{\theta}}\\
\end{equation}
The vertical component of the kick force is cancelled by gravity, while the remain horizontal component accelerates the trunk forward, thus we have:

\begin{equation}\label{eqn_3}
    M\ddot{x} = f\sin{\theta}\\
\end{equation}

By substituting \ref{eqn_2}, we get
\begin{align*}
        M\ddot{x} = \frac{Mg}{\cos{\theta}}\sin{\theta} = Mg\tan{\theta} = Mg\frac{x}{z}\\
\end{align*}
where, $x$, $z$ defines the CoM of the inverted pendulum. By rewriting above equation, we obtain a differential equation for the horizontal dynamics of the CoM:
\begin{equation}\label{eqn_4}
    \ddot{x} = \frac{g}{z}x\\
\end{equation}
During ground contact the planar locomotion of the CoM can be formulated by a system of two nonlinear differential equations \citep{blickhan1989spring}:
\begin{align*}
        \ddot{x}&=x\omega^2(\frac{r}{\sqrt{x^2+z^2}}-1)\numberthis \label{eqn4}\\
        \ddot{z}&=z\omega^2(\frac{r}{\sqrt{x^2+z^2}}-1)-g\numberthis \label{eqn5}
\end{align*}
where, $\omega=\sqrt{k/M}$ is the natural frequency of leg's spring-mass system, $k$ is the spring stiffness, and $M$ is the mass.
\subsubsection{Numerical Simulation of SLIP model}
\begin{figure}[h!]
	\centering
    \includegraphics[width=3.5cm]{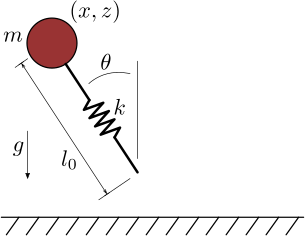}
	\caption{State machine and the transitions of an active IP model.}
	\label{fig:SLIP_2}
\end{figure}

As Fig. \ref{fig:SLIP_2} shown, the SLIP model is simplified as a point mass, $m$, on top of a massless, springy leg with rest length of $l_0$, and spring constant $k$. The configuration of the system is given by the $x, z$ position of the center of mass, and the length, $l$, and angle $\theta$ of the leg. The dynamics are modeled piece-wise as the combination of two stages: flight phase and stance phase.

During the flight phase, we use the state variables: $$x_f = [x,\dot{x},z,\dot{z}, \theta]^T$$
The corresponding flight dynamics are simply: $$F_f = [\dot{x},0,\dot{z},-g, \omega]^T = A_fx_f+B_f\omega+[0,0,0,-g,0]^T$$, where 
$$A_f = \begin{bmatrix}
0 & 1 & 0 & 0 &0\\
0 & 0 & 0 & 0 &0\\
0 & 0 & 0 & 1 &0\\
0 & 0 & 0 & 0 &0\\
0 & 0 & 0 & 0 &0\\
\end{bmatrix}	$$ and $$B_f = [0,0,0,0,1]^T	$$

To modeling the stance phase, we can either write the dynamics in polar coordinate or Cartesian coordinates system, with the foot anchored at the origin. Here we describe the state variables in Cartesian coordinates: $x_s = [\theta,\dot{\theta},l,\dot{l}]^T$. Plugging these into Lagrange yields the stance dynamics:
$$F_s = \begin{bmatrix}
\dot{\theta} \\
-\frac{1}{l}(2\dot{\theta}\dot{l}+g\cos{\theta}) \\
\dot{l} \\
-g\sin{\theta}+\dot{\theta}^2l+\frac{k}{m}(l_0-l) \\
\end{bmatrix}	+ 
\begin{bmatrix}
0&0 \\
0&\frac{1}{ml^2} \\
0&0 \\
\frac{k}{m}&0 \\
\end{bmatrix}
\begin{bmatrix}
0 \\
1.5\\
\end{bmatrix}$$ 
\begin{figure}[h!]
	\centering
    \includegraphics[width=14cm]{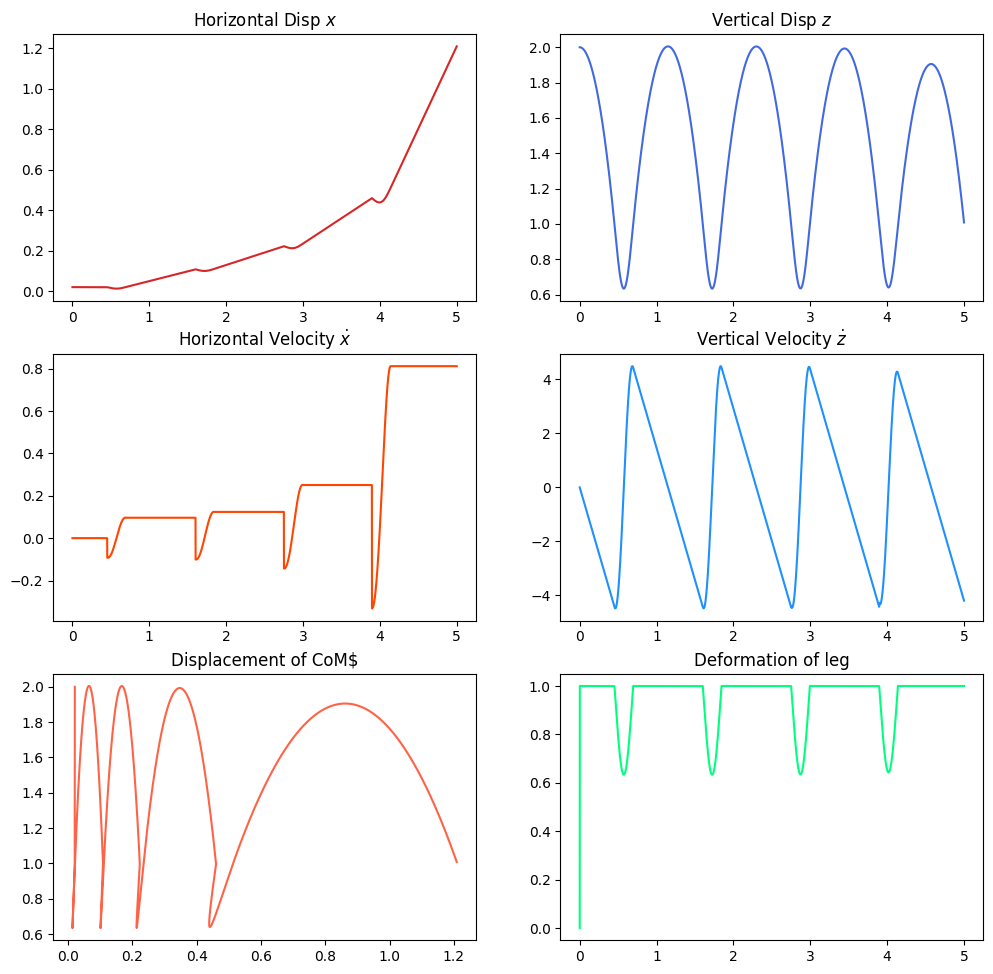}
	\caption{Numerical Simulation of SLIP model.}
	\label{fig:SLIP_num}
\end{figure}

We utilize the MATLAB code developed by \citep{brandt} to simulate a free-fall (without control input) of a SLIP model. In Fig. \ref{fig:SLIP_num}, we depicted six kinematic curves of the model: (a) horizontal displacement annotated as $x$ (b) vertical displacement annotated as $z$ (c) horizontal velocity annotated as $\dot{x}$ (d) vertical velocity annotated as $\dot{z}$ (e) displacement of CoM in x-z plane (f) axial displacement of the leg $l$. We can find that the SLIP model in free fall has a tendency to lean forward in motion due to the transformation of gravitational potential energy. By introducing the controlling of angle $\theta$, the SLIP model of continuous walking can be realized.

\subsubsection{Simulation of a 2D-Bipedal IP model}

\begin{figure}[h!]
	\centering
    \includegraphics[width=12cm]{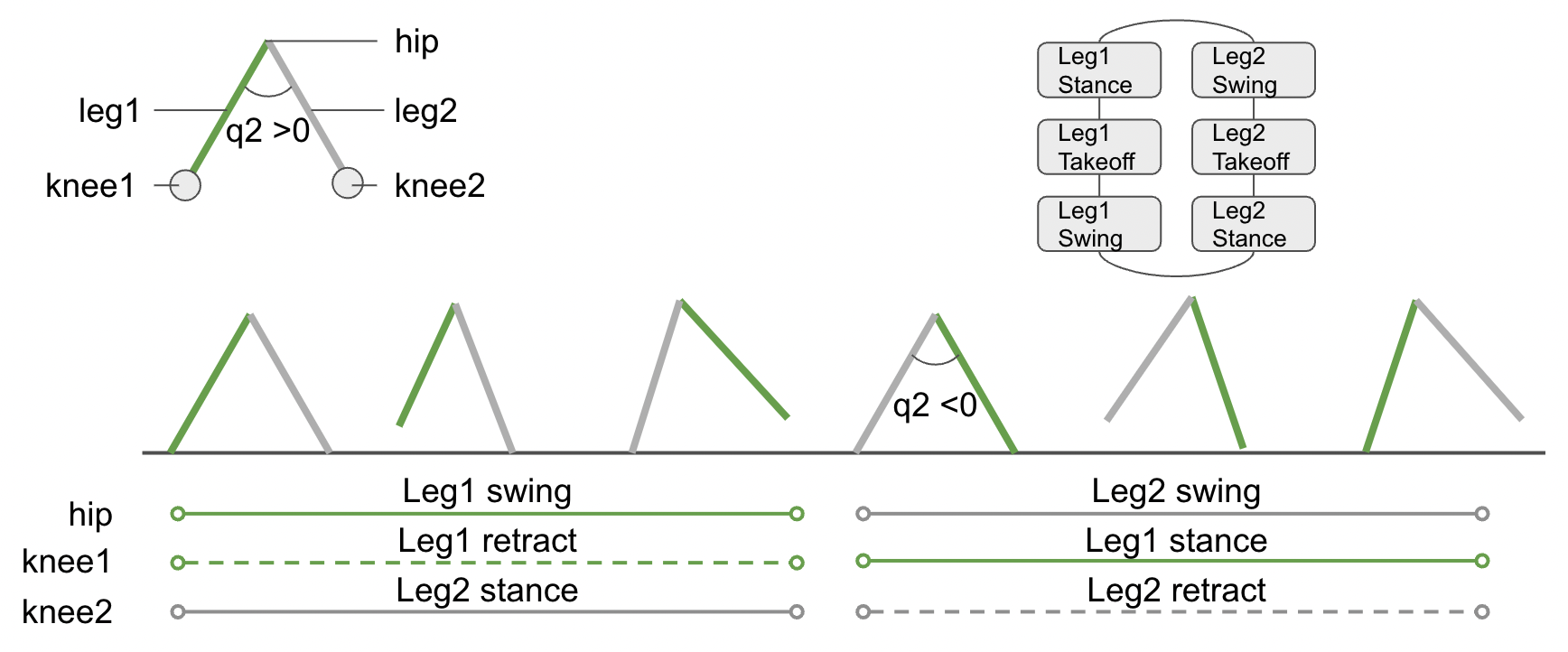}
	\caption{State machine and the transitions of an active IP model.}
	\label{fig:SLIP}
\end{figure}

We chose a physics engine called \textit{MuJoCo} \citep{todorov2012mujoco} to simulate the 2D-bipedal model. The C programming language was used in the environment of MuJoCo to create our moving simulation. As shown in Figure. \ref{fig:SLIP}, we will create a simple 2D leg model that can walk or run. To do so, we must first create a \codeword{xml} file in MuJoCo to represent our bipedal model. We started with the hip, knee, and foot joints because they are the most commonly used when walking. This is made up of several springs and hinges to represent the knees and hips, as well as two spheres to represent model's feet.

We assigned a gravity that allows the components to fall to the ground and maintain constant contact with the ground when it's moving, because people tend to lift themselves off the ground when moving forward. Another thing we'll make is the ground plane on which the simulation will walk. After that, we must separate the legs, or else the simulation will land straight and the bipedal model will fall backwards or forwards, resulting in no stable point to balance. We need to add speed to the model, which will be edited based on the state machine we created.

\clearpage


Another thing we did to make the model a moving/working simulation was to create the state machine. Figure 5:State is a simple way to understand the state machine that allows our bipedal model to move. The right leg takes off, then plants itself on the ground and remains in place. Later, the left leg swings and takes off, before landing on the ground. It then swings the right leg again and repeats the process for a set period of time.

To accomplish this, we needed to run three separate state machines at the same time to coordinate hip, knee, and foot movement. Following that, we had to set the restricted speed of the leg swing because if it is too fast, we will push the CoM backward following off, and if it is too slow, we will not walk. We also had to account for leaning forward, because if the model didn't lean forward and stood perfectly straight, we'd push ourselves back and fall backward. All body part numbers are linked to our MuJoCo model in order to be considered in the state machine.

\begin{figure}[h!]
	\centering
    \includegraphics[width=8cm]{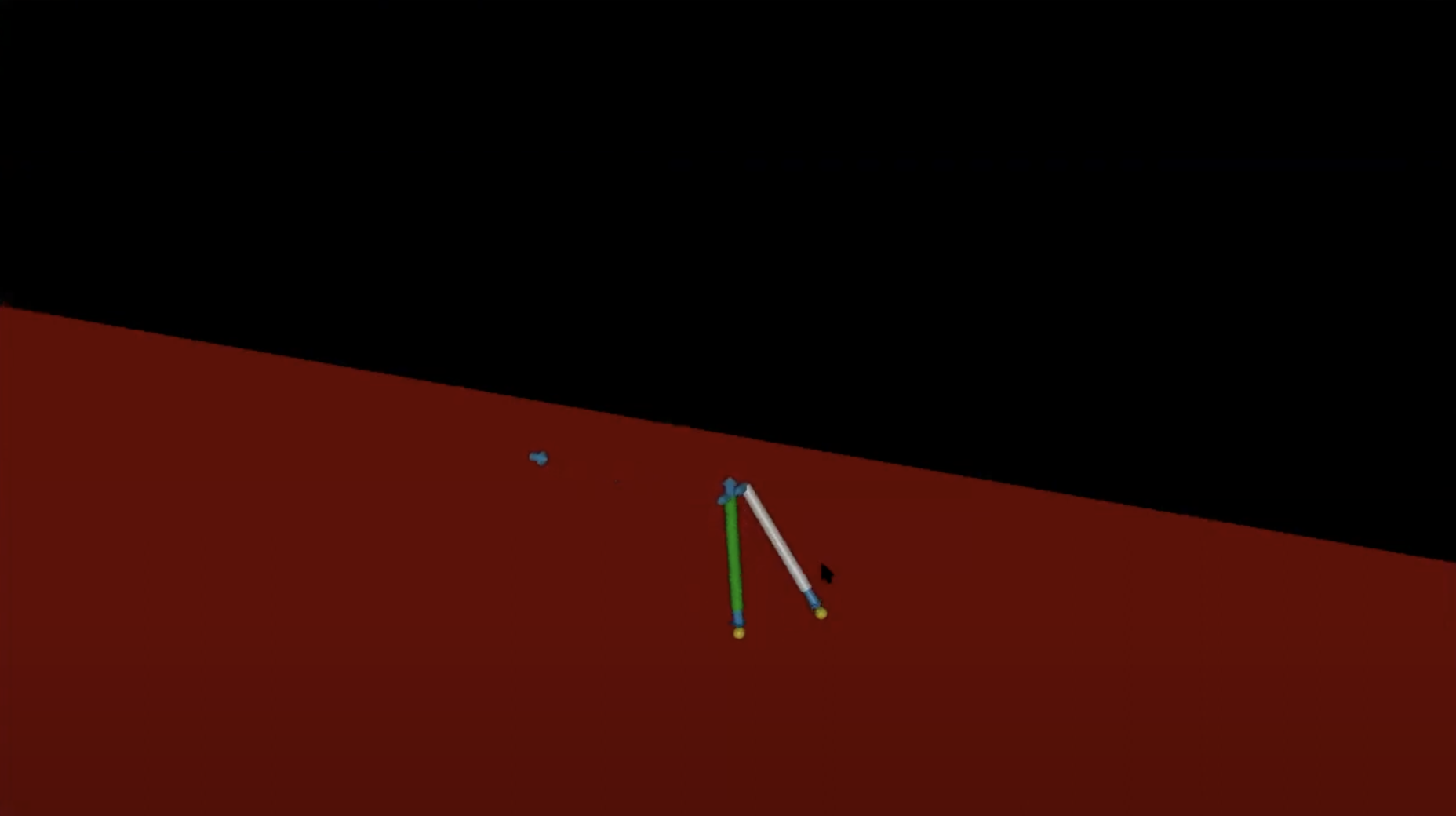}
	\caption{Mujoco Simulation.}
	\label{fig:2d simulation}
\end{figure}
The resulting simulation is shown in Figure. \ref{fig:2d simulation}, and the corresponding data is collected.

\subsection{IP-inspired Crawling Model}
\begin{figure}[h!]%
    \centering
    \subfloat[\centering Baby Crawl]{{\includegraphics[width=4cm]{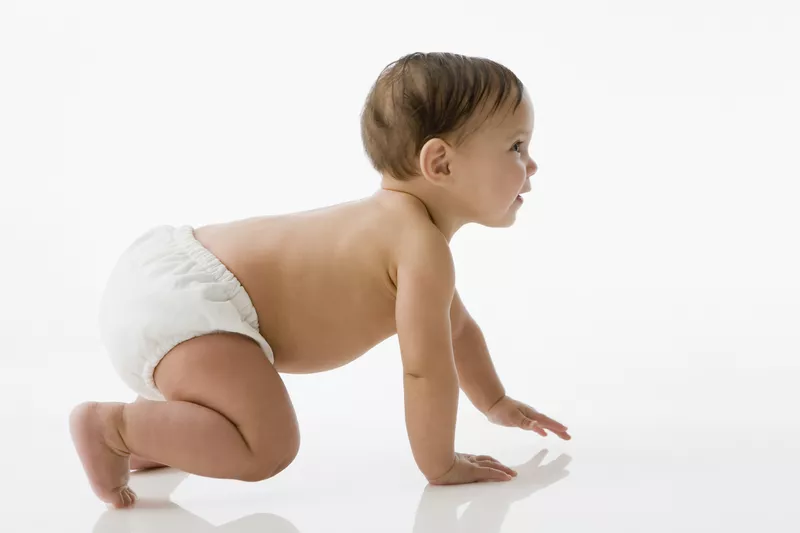} }}%
    \qquad
    \subfloat[\centering Combined inverse pendulums]{{\includegraphics[width=4cm]{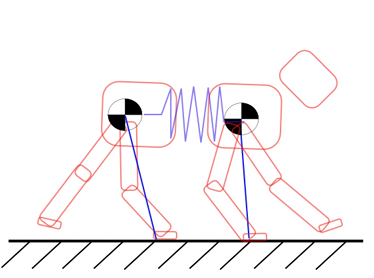} }} 
    \caption{Adaptation of the IP model for creating crawling model.}%
    \label{fig:Quadruped}%
\end{figure}

By observing the crawling movements of infants \footnote{See video: \url{https://t.ly/n7Q_}}, we found that infants tend to adopt a crawling mode in which the hind limbs sit and stand while the front trunk crawls as the main forward momentum, which looks very much like a frog but in this paper we used a quadrupedal walking model of mammals for simplicity. A quadruped model is shown in Figure. \ref{fig:Quadruped} (b), which consists two inverted pendulum model inspired by the single 2D-biped \citep{polet2019simple}.

\subsubsection{Free Body Diagram Analysis}

The model was constructed by a rigid trunk connects two point masses: fore mass $m_F$ and hind mass $m_H$. Each point mass has two massless limbs that can extend or contract. The front left and right legs are $leg_2$, $leg_1$, similarly, rear left and right legs are $leg_4$, $leg_3$. The swing angle of the limbs are defined as $q_{11}$, $q_{12}$, $q_{23}$, $q_{24}$, where the $i,j$ in $q_{ij}$ represents the front/rear group and $j^{th}$ leg. The swing angle between two legs in each group were defined as $q_1$ and $q_2$, and we have,
\begin{align*}
    q_1 = q_{11}+q_{12}\\
    q_2 = q_{23}+q_{24}\\
\end{align*}

\begin{figure}[h!]
	\centering
    \includegraphics[width=10cm]{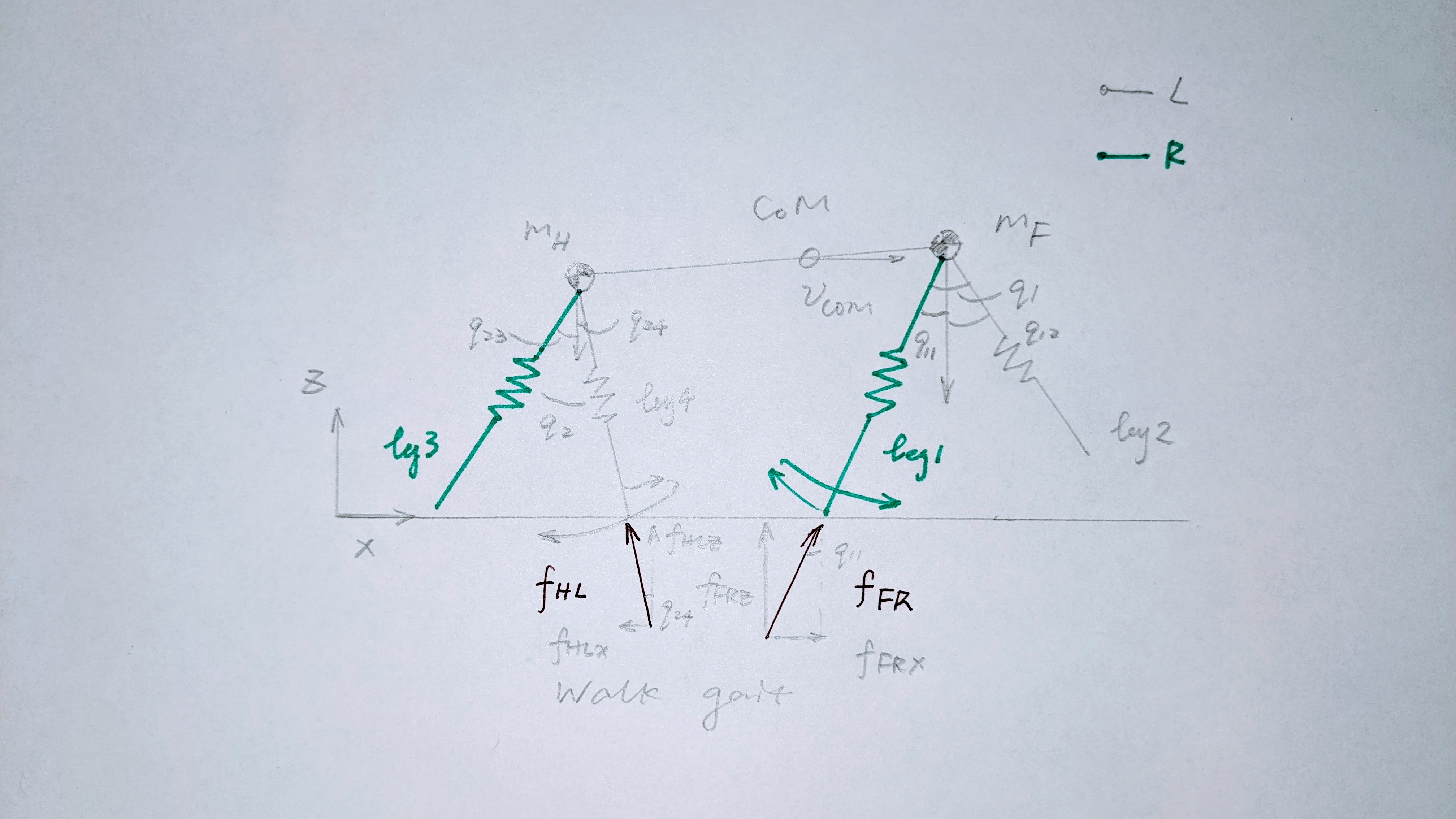}
	\caption{Free-body diagram of a simple quadruped model.}
	\label{fig:quad_fbd}
\end{figure}

Due to the complex behavior of a crawling baby \citep[Fig. 7]{righetti2015kinematic}, we consider the crawling model as the walking gait of a dog. We apply free body diagram analysis to the model as shown in Figure. \ref{fig:quad_fbd}. The contact force of $leg_1$ and $leg_4$ were $f_{FR}$ and $f_{HL}$ which drive the velocity of CoM as $v_{CoM}$ or $\ddot{x}_{CoM}$. The leg force project to the $x$ and $z$ axis, we have,
\begin{align*}
        f_{FRx} - f_{HLx}&= (m_H+m_F)\ddot{x} \numberthis\label{eqn6}\\
        f_{FRz} + f_{HLz}&= (m_H+m_F)g \numberthis \label{eqn7}
\end{align*}

apply the swing angles, we have,

\begin{align*}
        f_{FR}\sin{q_{11}} - f_{HLx}\cos{q_{24}}&= (m_H+m_F)\ddot{x} \numberthis\label{eqn8}\\
        f_{FRz}\cos{q_{11}} + f_{HLz}\sin{q_{24}}&= (m_H+m_F)g \numberthis \label{eqn9}
\end{align*}

\subsubsection{Modeling and Simulation in Mujoco}

We used a quadrupedal walking model of mammals to simplified the quadrupedal model. We imagine the front and rear legs of baby as two bipedal models. The mass $m_F,m_H$ of front and rear body are located at the hip joint of each. A solid bar link connect the front and rear legs by two hinge joints at the front/rear connection point. The connection point will be longer because we want to mimic how a human baby will look like and they are pretty long when sprawled straight. Once we made our connection between the front and the rear legs, we need to change the center of mass. If we have it perfectly in the middle, we will have issues with the model tipping over or falling over once it attempts to make its first crawling movement. To change the center of mass, we made the weight of the front legs more. Doing this allowed us to make a crawling motion for the baby where it will not fall over and lie there struggling to move. This showed us something important about how babies crawl when moving.  Babies will put more weight and force in the front half of their body and do a little leaning motion when moving forward very similar to how humans walk in a bipedal motion. Babies are basically pulling themselves at the same time while crawling which is being shown in our simulation mode. If they were to put no force or weight in the front of their own body, they risk just falling over on the floor every time since the force will be strong pushing them backwards and not being able to stand up properly to crawl anymore.

To create the control, or the movement of our quadrupedal model, we had to create state machines again to mimic the crawling movement. This is going to be very similar to how our bipedal moves. The right arm takes off, then plants itself on the ground and remains in place. Later, the left arm swings and takes off, before landing on the ground. It then swings the right arm again and repeats the process for a set period of time.

To accomplish this, we needed to run three separate state machines at the same time to coordinate arm and hand movement. Following that, we had to set the restricted speed of the arm swing because if it is too fast, we will push the CoM backward following off, and if it is too slow, the baby will fall face first flat on the ground. We also had to account for leaning forward, because if the model didn't lean forward and stood perfectly straight, we'd push ourselves back and fall backward. All body part numbers are linked to our MuJoCo model in order to be considered in the state machine. This is very similar to our bipedal model, however we will disregard the rear legs. We know that the babies have a little movement in rear leg when crawling, however due to the time constraints of the class being a quarter system, we decided that if we were able to get the model to crawl with just front legs, we would have succeeded in our goal of creating a crawling model.

\begin{figure}[h!]
	\centering
    \includegraphics[width=16cm]{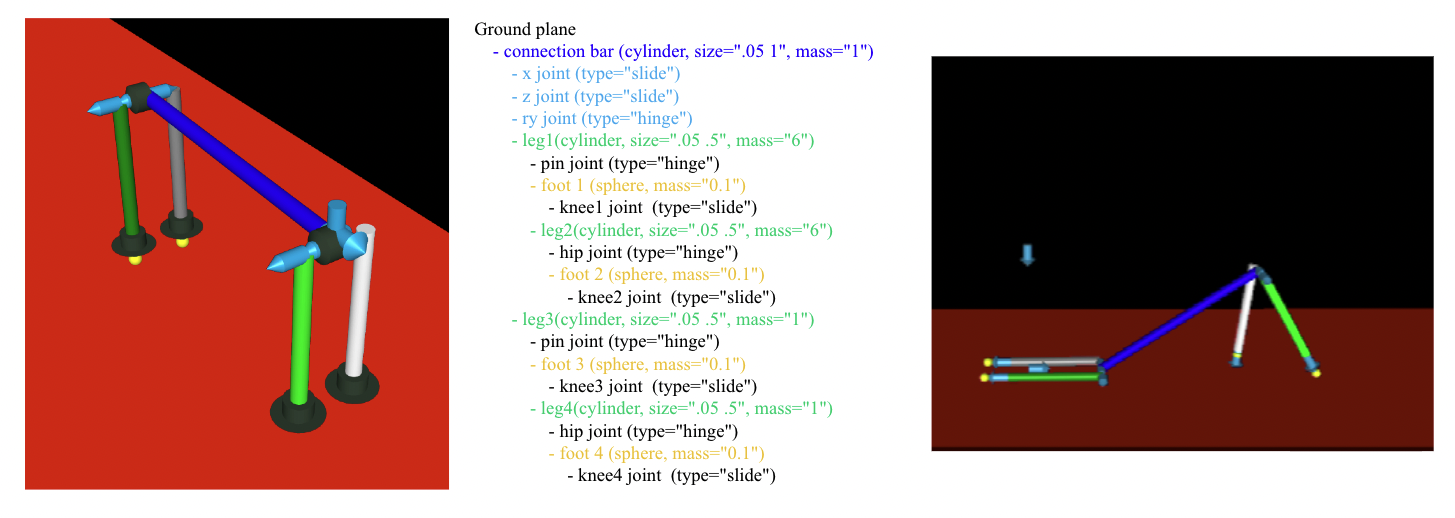}
	\caption{The hierarchical representation of our model and the simulation result of driven by front legs only.}
	\label{fig:final}
\end{figure}

The hierarchical representation of our model is shown in Fig \ref{fig:final}. We borrow the ratio of infant's skeleton key-points \citep{hesse2018learning,hesse2018computer} to build the MuJoCo model that matches the infant's body proportions.


\section{Discussion and Conclusion}
This project was interesting in it's own right. We chose to simulate a human baby walking, running, and crawling to understand this complex bio-inspired locomotion that we humans are capable of. As a group we were able to learn that humans when walking will naturally lean forward in order to move forward and also we never stand up perfectly 90 degrees when walking because it is neither efficient and if we swing to hard, we will risk pushing ourselves backward thus falling over in the process. Also when walking, we have a constant swinging motion of roughly the same speed every time or else we risk balancing issues. Another thing we observed is the fact that we hop a little bit each time when walking and never drag our feet on the ground which makes sense since dragging our feet on the ground would mean using way move energy than necessary or just not being able to move effectively from place to place. Crawling was interesting as well as the human baby follows a similar like pendulum motion when walking and a leaning forward motion, but we also learned that babies tend to put most of their weight in the front of the body where there arms are doing the movement to balance themselves and perpetuate forward when crawling. Our Matlab results showed that as we move further or made a bigger displacement with each movement in the leg, our horizontal velocity increased which made sense walking and running would have different horizontal velocities and running would cover more distance as opposed to walking. Vertical displacement and Vertical velocity stayed relatively the same which also made sense since we as humans do not want to be up in the air for an extended period of time, but also not be planted on the ground or else we would not be moving around efficiently. Meaning, if we stayed in the air too long, we would not be covering as much distance as we could have if we lightly hopped in the air, but if never left the ground we would have a hard time moving ones legs in turn making it hard to move from one place to another. We also analyzed the spring of a humans leg and it made sense that human's legs spring back and forth when up on the ground or making contact with the ground. We will be fully extended when we are at our peak vertical displacement and when we make contact with the ground, we will bend our legs a little bit shown in figure 4 deformation of leg graph mainly because we do not want to damage our legs when landing on the ground. Landing on the ground at 90 degrees every time from a certain distance will damage our legs over time causing arthritis in the future.As stated earlier, we would have incorporated movement in hind legs when crawling if we had more time, but due to the nature of the quarter system, we had to do what we could complete in the 10 weeks we were given and overall happy with the results that we were able to show as a group.

\section*{Acknowledgements}

We would like to than Mircea and the ECE 216 for helping us and providing feedback along the way to improve our project and paper.

\clearpage
\nocite{*}

\bibliographystyle{unsrt}
\bibliography{references}  

\begin{thebibliography}{10}

\bibitem{turing1950mind}
Alan~Mathison Turing.
\newblock Computing machinery and intelligence.
\newblock {\em Mind}, 59(236):433--460, 1950.

\bibitem{garcia2007evolution}
Elena Garcia, Maria~Antonia Jimenez, Pablo~Gonzalez De~Santos, and Manuel
  Armada.
\newblock The evolution of robotics research.
\newblock {\em IEEE Robotics \& Automation Magazine}, 14(1):90--103, 2007.

\bibitem{thrun2006stanley}
Sebastian Thrun, Mike Montemerlo, Hendrik Dahlkamp, David Stavens, Andrei Aron,
  James Diebel, Philip Fong, John Gale, Morgan Halpenny, Gabriel Hoffmann,
  et~al.
\newblock Stanley: The robot that won the darpa grand challenge.
\newblock {\em Journal of field Robotics}, 23(9):661--692, 2006.

\bibitem{krotkov2017darpa}
Eric Krotkov, Douglas Hackett, Larry Jackel, Michael Perschbacher, James
  Pippine, Jesse Strauss, Gill Pratt, and Christopher Orlowski.
\newblock The darpa robotics challenge finals: Results and perspectives.
\newblock {\em Journal of Field Robotics}, 34(2):229--240, 2017.

\bibitem{adolph2012you}
Karen~E Adolph, Whitney~G Cole, Meghana Komati, Jessie~S Garciaguirre, Daryaneh
  Badaly, Jesse~M Lingeman, Gladys~LY Chan, and Rachel~B Sotsky.
\newblock How do you learn to walk? thousands of steps and dozens of falls per
  day.
\newblock {\em Psychological science}, 23(11):1387--1394, 2012.

\bibitem{sharbafi2017bioinspired}
Maziar~Ahmad Sharbafi and Andr{\'e} Seyfarth.
\newblock {\em Bioinspired legged locomotion: models, concepts, control and
  applications}.
\newblock Butterworth-Heinemann, 2017.

\bibitem{blickhan1993similarity}
Reinhard Blickhan and RJ~Full.
\newblock Similarity in multilegged locomotion: bouncing like a monopode.
\newblock {\em Journal of Comparative Physiology A}, 173(5):509--517, 1993.

\bibitem{alexander1990three}
RMcN Alexander et~al.
\newblock Three uses for springs in legged locomotion.
\newblock {\em International Journal of Robotics Research}, 9(2):53--61, 1990.

\bibitem{blickhan1989spring}
Reinhard Blickhan.
\newblock The spring-mass model for running and hopping.
\newblock {\em Journal of biomechanics}, 22(11-12):1217--1227, 1989.

\bibitem{kajita2014introduction}
Shuuji Kajita, Hirohisa Hirukawa, Kensuke Harada, and Kazuhito Yokoi.
\newblock {\em Introduction to humanoid robotics}, volume 101.
\newblock Springer, 2014.

\bibitem{brandt}
Martin~Albertsen Brandt.
\newblock Simulation and control of pogo robot (spring loaded inverted
  pendulum) in matlab.

\bibitem{todorov2012mujoco}
Emanuel Todorov, Tom Erez, and Yuval Tassa.
\newblock Mujoco: A physics engine for model-based control.
\newblock In {\em 2012 IEEE/RSJ international conference on intelligent robots
  and systems}, pages 5026--5033. IEEE, 2012.

\bibitem{polet2019simple}
Delyle~T Polet and John~EA Bertram.
\newblock A simple model of a quadruped discovers single-foot walking and
  trotting as energy optimal strategies.
\newblock {\em BioRxiv}, page 580779, 2019.

\bibitem{righetti2015kinematic}
Ludovic Righetti, Anna Nyl{\'e}n, Kerstin Rosander, and Auke~Jan Ijspeert.
\newblock Kinematic and gait similarities between crawling human infants and
  other quadruped mammals.
\newblock {\em Frontiers in neurology}, 6:17, 2015.

\bibitem{hesse2018learning}
Nikolas Hesse, Sergi Pujades, Javier Romero, Michael~J. Black, Christoph
  Bodensteiner, Michael Arens, Ulrich~G. Hofmann, Uta Tacke, Mijna
  Hadders-Algra, Raphael Weinberger, Wolfgang M\"uller-Felber, and A.~Sebastian
  Schroeder.
\newblock Learning an infant body model from {RGB-D} data for accurate full
  body motion analysis.
\newblock In {\em International Conference on Medical Image Computing and
  Computer-Assisted Intervention (MICCAI)}. Springer, 2018.

\bibitem{hesse2018computer}
Nikolas Hesse, Christoph Bodensteiner, Michael Arens, Ulrich~G. Hofmann,
  Raphael Weinberger, and A.~Sebastian Schroeder.
\newblock Computer vision for medical infant motion analysis: State of the art
  and {RGB-D} data set.
\newblock In {\em Computer Vision - ECCV 2018 Workshops}. Springer
  International Publishing, 2018.

\bibitem{cangelosi2015developmental}
Angelo Cangelosi and Matthew Schlesinger.
\newblock {\em Developmental robotics: From babies to robots}.
\newblock MIT press, 2015.

\bibitem{mordatch2012discovery}
Igor Mordatch, Emanuel Todorov, and Zoran Popovi{\'c}.
\newblock Discovery of complex behaviors through contact-invariant
  optimization.
\newblock {\em ACM Transactions on Graphics (TOG)}, 31(4):1--8, 2012.

\bibitem{sinnet20092d}
Ryan~W Sinnet and Aaron~D Ames.
\newblock 2d bipedal walking with knees and feet: A hybrid control approach.
\newblock In {\em Proceedings of the 48h IEEE Conference on Decision and
  Control (CDC) held jointly with 2009 28th Chinese Control Conference}, pages
  3200--3207. IEEE, 2009.

\bibitem{reddy2011modeling}
NS~Reddy, Ranjit Ray, and SN~Shome.
\newblock Modeling and simulation of a jumping frog robot.
\newblock In {\em 2011 IEEE international conference on mechatronics and
  automation}, pages 1264--1268. IEEE, 2011.

\bibitem{wisse2006design}
Martijn Wisse and Jan~van Frankenhuyzen.
\newblock Design and construction of mike; a 2-d autonomous biped based on
  passive dynamic walking.
\newblock In {\em Adaptive motion of animals and machines}, pages 143--154.
  Springer, 2006.

\bibitem{sharbafi_2017}
Maziar~A. Sharbafi.
\newblock Template models, Nov 2017.

\bibitem{mcmahon2018yeast}
Conor McMahon, Alexander~S Baier, Roberta Pascolutti, Marcin Wegrecki, Sanduo
  Zheng, Janice~X Ong, Sarah~C Erlandson, Daniel Hilger, S{\o}ren~GF Rasmussen,
  Aaron~M Ring, et~al.
\newblock Yeast surface display platform for rapid discovery of
  conformationally selective nanobodies.
\newblock {\em Nature structural \& molecular biology}, 25(3):289--296, 2018.

\end{thebibliography}






\end{document}